\documentclass[letterpaper, 10 pt, conference]{ieeeconf}  % Comment this line out if you need a4paper
\usepackage{amsmath, amssymb}
\usepackage{graphicx}
\usepackage{float} 
\usepackage{booktabs}
\usepackage{cite}
\usepackage{balance}
\usepackage{caption}
\usepackage{tcolorbox}
\usepackage{algorithm}
\usepackage{algpseudocode}
\usepackage{ulem}
\usepackage{color}
\usepackage{bm} 
\usepackage{soul}
\usepackage{threeparttable}

\makeatletter

\newcommand{\Rmnum}[1]{\expandafter\@slowromancap\romannumeral #1@}
\makeatother

\usepackage{CJKutf8}

\IEEEoverridecommandlockouts                              % This command is only needed if 
\title{\LARGE \bf
CS3D: An Efficient Facial Expression Recognition via Event Vision}

\author{Zhe Wang, Qijin Song, Yucen Peng, and Weibang Bai$^\ast$ %<-this }stops a space
\thanks{
This work is supported by the Shanghai Pujiang Program under grant 23PJ1408500, by the Shanghai Frontiers Science Center of Human-centered Artificial Intelligence (ShangHAI), MoE Key Laboratory of Intelligent Perception and Human-Machine Collaboration (KLIP-HuMaCo). The experiments of this work were supported by the Core Facility Platform of Computer Science and Communication, SIST, ShanghaiTech University.
Corresponding author: Weibang Bai \textit{(wbbai@shanghaitech.edu.cn)}.}
\thanks{Zhe Wang, Qijin Song, Yucen Peng, and Weibang Bai are with the ShanghaiTech Automation and Robotics (STAR) Center, School of Information Science and Technology, ShanghaiTech University, Shanghai, 201210, China.}
}

\begin{document}
\maketitle
\thispagestyle{empty}
\pagestyle{empty}

%%%%%%%%%%%%%%%%%%%%%%%%%%%%%%%%%%%%%%%%%%%%%%%%%%%%%%%%%%%%%%%%%%%%%%%%%%%%%%%%
\begin{abstract}

% With the development of the robotics industry, low-cost service robots will become a future trend. Therefore, accurately recognizing facial expression changes using edge devices such as the Jetson Nano and Raspberry Pi is crucial for enabling effective human-robot interaction in service robots. Event cameras surpass RGB cameras in capturing facial expression changes due to their high temporal resolution, low latency, and efficient processing. Traditional methods typically employ large deep learning models for facial expression analysis, and these methods are particularly energy-intensive, making them difficult to deploy on edge devices and consequently increasing robot costs. To address this task, we propose the CS3D model, which decomposes the C3D model to reduce computational complexity and lower energy consumption for deployment on edge devices. In addition, by utilizing soft spiking neurons and a spatio-temporal attention mechanism, the task gains an improved ability to retain information, thus improving the accuracy of facial expression recognition. Experimental results indicate that our proposed method attains higher accuracy on multiple datasets compared to architectures such as the RNN, Transformer, and C3D model, while the energy consumption of the CS3D model is just 21. 97\% of the original C3D model required on the same device.

Responsive and accurate facial expression recognition is crucial to human-robot interaction for daily service robots. Nowadays, event cameras are becoming more widely adopted as they surpass RGB cameras in capturing facial expression changes due to their high temporal resolution, low latency, computational efficiency, and robustness in low-light conditions. 
% adaptive to insufficient lighting environments. 
Despite these advantages, event-based approaches still encounter practical challenges, particularly in adopting mainstream deep learning models. Traditional deep learning methods for facial expression analysis are energy-intensive, making them difficult to deploy on edge computing devices and thereby increasing costs, especially for high-frequency, dynamic, event vision-based approaches. To address this challenging issue, we proposed the CS3D framework by decomposing the Convolutional 3D method to reduce the computational complexity and energy consumption. Additionally, by utilizing soft spiking neurons and a spatial-temporal attention mechanism, the ability to retain information is enhanced, thus improving the accuracy of facial expression detection. Experimental results indicate that our proposed CS3D method attains higher accuracy on multiple datasets compared to architectures such as the RNN, Transformer, and C3D, while the energy consumption of the CS3D method is just 21.97\% of the original C3D required on the same device.
\end{abstract}

%%%%%%%%%%%%%%%%%%%%%%%%%%%%%%%%%%%%%%%%%%%%%%%%%%%%%%%%%%%%%%%%%%%%%%%%%%%%%%%%
\section{INTRODUCTION}
With the rapid development of service robots across diverse domains, such as healthcare, education, and domestic assistance, etc., real-time facial expression recognition (FER) has emerged as a cornerstone for enabling natural and empathetic human-robot interaction\cite{faria2017affective,yang2008facial,rawal2022facial}. 
It is widely acknowledged that FER can generally be performed using different types of cameras \cite{huang2019facial}.

Conventional RGB cameras, the default sensors in most robotic systems, however, face inherent limitations in capturing transient facial muscle movements
% , such as the rapid activation of Action Units (AUs) defined by the Facial Action Coding System (FACS)
\cite{ekman1978facial}.  
% Due to muscle activation, the human face constantly exhibits subtle yet rapid movements that frequently occur abruptly and unpredictably. 
The facial subtle yet rapid movements, often lasting less than 500 milliseconds, are critical for decoding underlying emotions but are frequently obscured by RGB cameras’ low temporal resolution\cite{burrows2008facial}. 
High-frame-rate cameras\cite{zhao2023more} can improve expression recognition accuracy rate, while they generate a vast amount of frame data, leading to significant computational overhead and high energy consumption. 
As a result, it is not practical for large-scale applications and is not widely adopted in FER.
% This not only increases the operational cost of service robots, limiting their large-scale adoption, but also poses challenges to environmental sustainability.
%
% An event camera is a bio-inspired sensor that does not generate synchronized frame streams but instead produces asynchronous events when the illumination of a single-pixel changes. Its advantage lies in the extremely high speed of event occurrence, with a time resolution reaching the microsecond level, allowing it to capture rapid facial muscle changes while generating asynchronous events. Meanwhile, event cameras use asynchronous driving mechanism, collecting data only when illumination changes, reducing data redundancy and computation, and significantly lowering energy consumption.
%
Event camera \cite{gallego2020event} is a kind of bio-inspired sensor producing asynchronous events when the illumination of a single pixel changes, which is advantageous in extremely high-speed event occurrence, such as rapid facial muscle changes. 
Therefore, we adopt event cameras as the vision sensor for FER in this work.

Nevertheless, due to the high-frequency dynamic nature of event vision, current event-camera-based FER methods still face significant challenges of high computational complexity and suboptimal accuracy. To address the challenges, we propose CS3D, which combines soft spiking neurons \cite{zenke2018superspike}, factorized 3D convolutions \cite{wang2017factorized}, and spatial-temporal joint attention \cite{zhu2024tcja} to enhance efficiency and accuracy in facial expression recognition. 
Firstly, we utilize the V2E converter \cite{hu2021v2e} to preprocess the existing FER video datasets, to generate a sufficient event stream dataset. Subsequently, the preprocessed datasets are used to train the proposed CS3D architecture. Finally, we conduct a series of experiments to evaluate the proposed model and compare it with conventional algorithms in terms of recognition accuracy and energy consumption.

The main contributions of this work are as follows:
% \begin{itemize}
%     \item 
%     % By introducing a soft-thresholding and sigmoid surrogate gradient mechanism, the implementation of soft spiking neurons (SSNs) is proposed.
%     We propose an implementation of soft spiking neurons (SSNs) by introducing a soft-thresholding and sigmoid surrogate gradient mechanism. 
%     This mechanism better preserves input information and method temporal features during forward propagation. It also stabilizes backward gradient flow, improving the training efficiency and accuracy rate of deep SSNs. 
%     \item We propose a factorized 3D convolution module that decomposes 3D convolutions into separate temporal and spatial convolutions, reducing computational costs while improving efficiency in capturing dynamic expression changes and facial details.
%     \item We propose a spatio-temporal joint attention mechanism that integrates temporal and spatial attention modules, automatically focusing on the most critical time frames and facial regions 
%     % (such as the eyes and mouth) 
%     during expression changes, thereby enhancing feature representation and recognition accuracy.
% \end{itemize}

\begin{itemize}
    \item 
    We propose CS3D, a compact spatial-temporal 3D network architecture that integrates factorized 3D convolutions with a spatial-temporal joint attention mechanism.
    % dual-branch attention mechanism. 
    By jointly modeling temporal and spatial dependencies through temporal and spatial attention models, CS3D enhances discriminative feature representation while maintaining computational efficiency.%enabling robust performance in event-based facial expression recognition.
    \item 
    A factorized 3D convolution module is designed to improve 3D Convolutional Networks (C3D) by integrating factorized convolutions, soft spiking neuron (SSN), and residual connections. This module can help to achieve reduced computational complexity while enhancing temporal and directional feature extraction, making it well-suited for processing event-based data.
    \item 
    We conducted a series of experiments to verify our proposed CS3D framework by comparing energy consumption on different devices, evaluating accuracy on event-converted datasets, and testing expression recognition under sufficient and insufficient lighting conditions with real event camera data.
\end{itemize}

\section{RELATED WORK}
\subsection{RGB Camera-based FER} 
Current RGB Camera-based FER methods can be divided into two categories: static (frame-based) methods and dynamic (sequence-based) methods. 
For frame-based methods, FER can be performed using single images, and the current primary approaches include CNN-based \cite{liu2016facial,xie2017facial,shin2016baseline} and Transformer-based \cite{zhao2021former,huang2021facial} methods. 
For sequence-based methods, FER can be performed with temporal information encoded using consecutive frames or with overall information captured by aggregating key individual frames from video sequences \cite{yan2025observe}. These methods rely primarily on deep network architectures, such as 3D CNN \cite{thuseethan2023deep3dcann}, recurrent neural networks (RNN)\cite{zhang2021novel}, and Transformers\cite{liu2023expression}.
% Furthermore, RGB camera-based methods encounter the intrinsic challenge of dark or insufficient lighting environments, even being incapable in this case.
%
Furthermore, RGB camera-based methods struggle in dark, insufficient, or extreme lighting conditions, where their performance significantly degrades or even fails completely\cite{xie2025joint,li2025mask}.
% RGB cameras exhibit fundamental sensitivity constraints in low-light environments

% There are multiple implementations of sequence-based methods. 
% on one hand, temporal information can be encoded using consecutive frames. These methods primarily rely on deep network architectures, such as 3D CNN \cite{thuseethan2023deep3dcann}, recurrent neural networks (RNN)\cite{zhang2021novel}, and Transformers\cite{liu2023expression}; on the other hand, overall information can be captured by aggregating individual frames, for example, by adopting an “onset-apex-offset” \cite{yao2018texture} or “onset-apex” \cite{poux2021dynamic} schemes.

\subsection{Event Camera-based FER} 
Event cameras have demonstrated outstanding performance in various computer vision tasks, such as hand pose estimation \cite{rudnev2021eventhands}, object recognition \cite{kim2021n}, human pose estimation \cite{goyal2023moveenet}, as well as FER. For example, Barchid et al. \cite{barchid2023spiking} proposed a novel spiking neural network architecture called Spiking-FER. Berlincioni et al. \cite{berlincioni2023neuromorphic} used the traditional C3D algorithm to recognize facial expressions captured by an event camera. Becattini et al. \cite{becattini2024neuromorphic} proposed a neuromorphic facial analysis method based on cross-modal supervision. By constructing the FACEMORPHIC multimodal dataset and leveraging the temporal synchronization between RGB videos and event streams, they used 3D facial shape coefficients from RGB videos as supervision signals to train a facial action unit classifier on event camera data. Xiao et al. \cite{xiao2024estme} introduced the Event-Enhanced Motion Extractor model and the Event-Guided Attention model to leverage the high temporal resolution of event signals captured by event cameras. 
% These modules enhance the motion features of facial expressions and guide the network to focus on key regions of facial expressions, thereby improving facial expression recognition performance. 
Those aforementioned methods demonstrate strong capabilities in achieving accurate facial emotion detection. However, energy consumption still remains problematic, making it difficult to deploy on edge computing devices and resulting in high costs for service robots.

\subsection{Spiking Neural Networks} 
Recently, learning algorithms derived from the backpropagation algorithm, such as surrogate gradient learning \cite{li2021differentiable}, have enabled the training of deep spiking neural network (SNN) architectures by addressing the non-differentiability issue of spiking neurons. In recent years, SNNs have been widely applied to computer vision tasks, such as video classification \cite{wu2018spiking}, action recognition \cite{liu2021event}, and expression recognition \cite{fu2012spiking} due to their ability to capture temporal dynamic features. Although SNNs theoretically offer several advantages, they still face various challenges. Firstly, the hard thresholding activation function hinders gradient propagation during backpropagation, thereby limiting the optimization of deep networks. Secondly, traditional SNNs rely solely on spike signals for information transmission, which can lead to the loss of continuous features during propagation and ultimately weaken their feature representation capabilities.

\subsection{3D Convolutional Networks (C3D)} 
C3D is a neural network that leverages 3D convolutions to jointly model spatial and temporal features, widely used in video understanding tasks\cite{tran2015learning}. C3D network has inspired numerous video analysis and recognition studies to design more effective spatial-temporal feature modeling approaches. Lea et al.\cite{lea2017temporal} mentioned that C3D can be used to extract spatial-temporal frame-level features as input to the temporal convolutional networks, enhancing its temporal modeling capability in action recognition. Duan et al.\cite{duan2022revisiting} adopted the classic C3D network as one of the backbone models in their PoseConv3D framework to process 3D pose heatmap volumes and evaluate its spatial-temporal modeling capability for skeleton-based action recognition. 
%Kang et al.\cite{kang2024asf} \redcolor{introduced C3D with 3D convolutions}, which inspired \redcolor{SSFF's} multiscale feature fusion design. 

\section{METHOD}
\begin{figure*}[h!]
    \centering
    \includegraphics[width=0.95\linewidth]{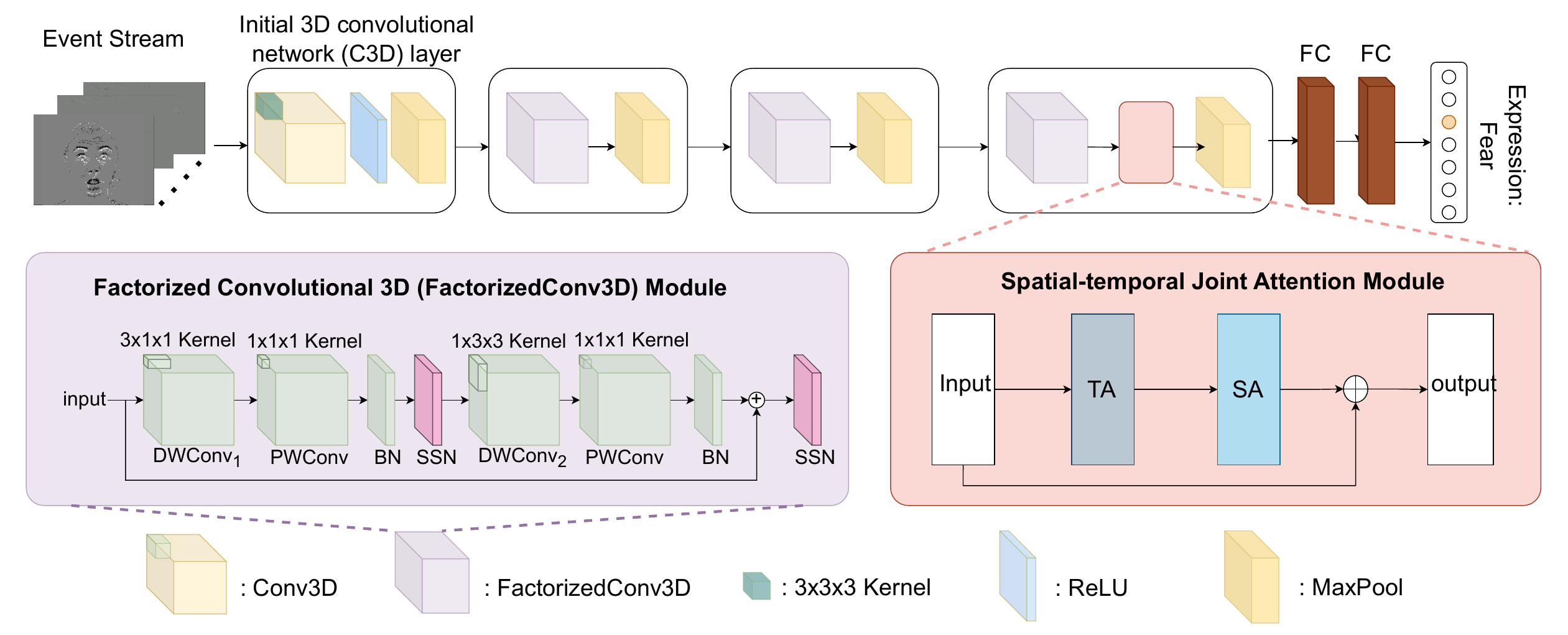}
    \caption{Overview of the proposed CS3D framework. The upper row describes the overall architecture of the CS3D. 
    The bottom row illustrates the FactorizedConv3D module and the spatial-temporal joint attention module integrated in the framework. FactorizedConv3D decomposes standard 3D convolutions to reduce the number of parameters and lower the time and space costs of model operation. The spatial-temporal joint attention module integrates temporal and spatial attention, enhancing the model’s ability to capture critical temporal and spatial information in the event stream.
    % , with TA denoting the temporal attention module and SA denoting the spatial attention module.
    % The bottom row illustrates framework details about the FactorizedConv3D module and the spatial-temporal joint attention module, where TA denotes the temporal attention module and SA denotes the spatial attention module, respectively.
    }
    \label{fig:overview}
    \vspace{-4mm}
\end{figure*}

The proposed CS3D framework is shown in Fig.\ref{fig:overview}. First, the event stream is fed into a FactorizedConv3D model, which reduces computational complexity by factorizing the convolutional kernels and extracting initial spatial-temporal features. Next, a Multi-Pool layer further enriches the feature representation. 
% Based on the above, the network introduces both Temporal Attention (TA) and Spatial Attention (SA) module, the former focusing on salient regions in the image space, and the latter highlighting keyframes in the temporal sequence. 
Finally, the Combined Attention Module integrates Temporal Attention (TA) and Spatial Attention (SA) to obtain a more comprehensive spatial-temporal feature representation, which is then passed through a fully connected or classification layer to produce the final prediction. This pipeline enables more effective FER from event streams.
% \redcolor{( is there any methods intoduced in the paragarph is proposed by us, if any, just higlight and use "propose".)}

\subsection{Soft Spiking Neuron} 
% Conventional SNNs use hard thresholding activation, producing output only when input exceeds a threshold. This causes the vanishing gradient problem, limiting their deep learning performance. 
Conventional SNNs use hard thresholding activation, where a neuron emits a binary spike only when its membrane potential exceeds a predefined threshold. This causes the vanishing gradient problem, limiting their deep learning performance. To address this issue, a Soft Spiking Neuron (SSN) is proposed, which approximates ReLU in the forward propagation through a Soft-Thresholding mechanism. At the same time, during backpropagation, the sigmoid surrogate gradient is used to improve the gradient flow, enabling the SNN to maintain biological interpretability while improving training stability and accuracy.

In the proposed SSN structure, the output function of the spiking neuron is defined as:

\begin{equation}
    f(x) = 
\begin{cases}
x, & x > \theta, \\
0, & x \le \theta.
\end{cases}
\end{equation}
where $\theta$ is the threshold of the neuron and $x$ is the input signal. When the input signal exceeds $\theta$, the neuron no longer only outputs discrete spikes but can transmit continuous information. This allows the network to retain more feature information and improve computational efficiency.

To address the discontinuity of soft-thresholding activation at the threshold, the sigmoid surrogate gradient is introduced during backpropagation as follows:
\begin{equation}
f'(x) = \sigma(\beta (x - \theta))
\end{equation}
where $\sigma(x)$ is the sigmoid function and $\beta$ controls the steepness of the curve. When $\beta \to \infty$, the sigmoid function becomes a step function and when $\beta$ takes smaller values, the gradient becomes smoother.

\subsection{Factorized 3D Convolution Module}
% The C3D algorithm is a deep learning method based on 3D convolutional neural networks, specifically designed for processing video data \cite{tran2015learning}. By applying 3D convolutional kernels on consecutive frames of a video, it can capture both spatial and temporal information, thereby extracting dynamic features from the video. In facial expression recognition, this method not only identifies spatial features in individual frames but also captures the temporal variations of facial muscle movements in the video, enabling the inference of changes in facial expressions over time.

% However, the standard 3D convolution is computationally expensive, with many parameters, which limits its application in resource-constrained scenarios. In service robots, computational devices typically rely on embedded systems or low-energy processors, which have strict limitations in terms of computational power, storage space, and energy consumption. Therefore, optimizing and simplifying 3D convolution becomes essential to enable real-time task processing, improve system response speed, and reduce energy consumption, ensuring that the robot can still efficiently perform complex visual perception and interaction tasks with limited hardware resources. 
The C3D architecture, originally proposed for RGB video analysis~\cite{tran2015learning}, learns spatial-temporal features with 3D convolutions; in this work, we adapt it to event-stream data. Standard 3D convolution is computationally expensive and high in parameter count, restricting its feasibility on the edge computing devices of service robots.
% To address this problem, factorized 3D convolution is proposed. Its structure is shown in Fig. \ref{fig:C3DS}. 
% \begin{figure}[htp]  
%     \centering
%     \includegraphics[width=1\linewidth]{Figures/FactorizedConv3D.pdf}
%     \caption{Factorized 3D Convolution blocks with SSN \redcolor{XXX?}}
%     \label{fig:C3DS}
%     \vspace{-4mm}
% \end{figure}
Factorized 3D convolution module contains two depth-wise convolution (DWConv) layers, two identical point-wise convolution (PWConv) layers, two batch normalization (BN) layers and two SSN layers. $\text{DWConv}_1$ uses $3\times 1 \times 1$ kernel size and  $\text{DWConv}_2$ uses $1\times 3 \times 3$ one. This difference makes DWConv layers respectively along the temporal dimension and the spatial dimension. The PWConv layers are applied to achieve information fusion between channels. Also, the network structure leverages the residual connection to ensure efficient gradient propagation. Through decomposing 3D convolutional module into temporal and spatial convolutions, factorized 3D convolution module reduces the number of parameters and lowers the time and space costs of model operation.

\subsection{Spatial-temporal Joint Attention Module}
To better focus on key facial regions, we introduce a spatial-temporal joint attention mechanism, which enhances the model’s ability to capture critical temporal and spatial information.
%This module consists of TA\cite{hu2018squeeze} and SA\cite{chen2024ehoa}, as shown in Fig. \ref{fig:Temporal Attention} and Fig. \ref{fig:Spatial Attention}.

% SSN is employed as the activation function in the TA module. The 1D convolution offers lower computational complexity and fewer parameters compared to traditional 2D or 3D convolutions, effectively reducing resource consumption and improving processing speed. Furthermore, max pooling and average pooling are combined for attention computation to capture more robust spatio-temporal features.

Temporal Attention (TA):
The TA module is an attention mechanism that assigns adaptive weights to different timesteps in a sequence to emphasize keyframes and suppress irrelevant frames. This design enables the CS3D architecture to emphasize keyframes by assigning adaptive temporal weights while suppressing redundant frames through dynamic feature reweighting.

% The TA module adaptively adjusts feature weights across timesteps, enhancing keyframe representations while suppressing redundant frame through dynamic feature reweighting.
\begin{figure}[htp]
    \centering
    \includegraphics[width=0.95\linewidth]{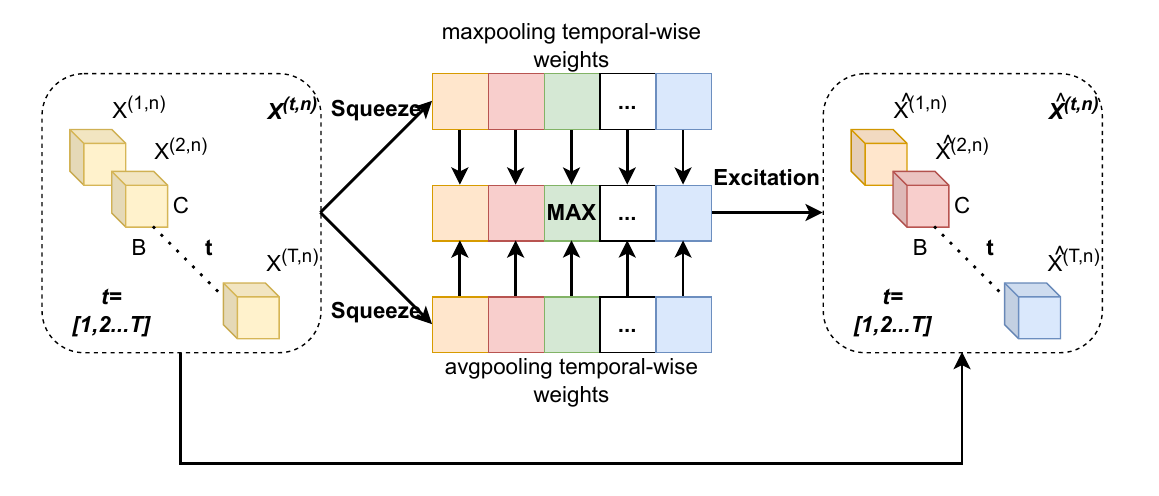}
    \caption{Temporal Attention \cite{hu2018squeeze,chen2024ehoa}}
    \label{fig:Temporal Attention}
    % \vspace{-5mm}
\end{figure}

The adopted TA module structure \cite{hu2018squeeze,chen2024ehoa} is shown in Fig. \ref{fig:Temporal Attention}. The spatial global average and max pooling extract two temporal-wise weight sets, which are processed by shared convolution and activation, fused via element-wise max to produce attention weights, and applied through residual connection to highlight key frames and suppress redundancy for improved temporal modeling.
The detailed implementation of the TA module is summarized in Algorithm \ref{alg:attention_residual}.
\begin{algorithm}[htp]
    \small
    \caption{Temporal Attention Module}
    \label{alg:attention_residual}
    \begin{algorithmic}[1] % [1] 使代码自动编号
        \State \textbf{Input:} $X \in \mathbb{R}^{B \times C \times T \times H \times W}$
        \State \textbf{Output:} $\hat{X} \in \mathbb{R}^{B \times C \times T \times H \times W}$
        \State $z_{\text{avg}} \leftarrow \frac{1}{H \cdot W} \sum_{h=1}^{H} \sum_{w=1}^{W} X_{(b,c,t,h,w)}$
        \State $z_{\text{max}} \leftarrow \max\limits_{1 \leq h \leq H, 1 \leq w \leq W} X_{(b,c,t,h,w)}$
        \State $S_t \gets \sigma(\text{Conv}_2(\varphi(\text{Conv}_1(z_{\text{avg}}))))$
        \State $S_s \gets \sigma(\text{Conv}_2(\varphi(\text{Conv}_1(z_{\text{max}}))))$
        \State $S \gets \max(S_t, S_s)$
        \State $\hat{X} \gets X \cdot S + X $
        \State \Return $\hat{X}$
    \end{algorithmic}
\end{algorithm}

Spatial Attention (SA):
The SA module introduces an attention mechanism to automatically enhance discriminative regions in input feature maps, guided by inter-channel statistical patterns and residual learning principles. This design enables the CS3D architecture to focus on critical areas by assigning adaptive weights while suppressing background interference through gradient-propagatable feature reweighting.
\begin{figure}[htp]
    \centering
    \includegraphics[width=0.92\linewidth]{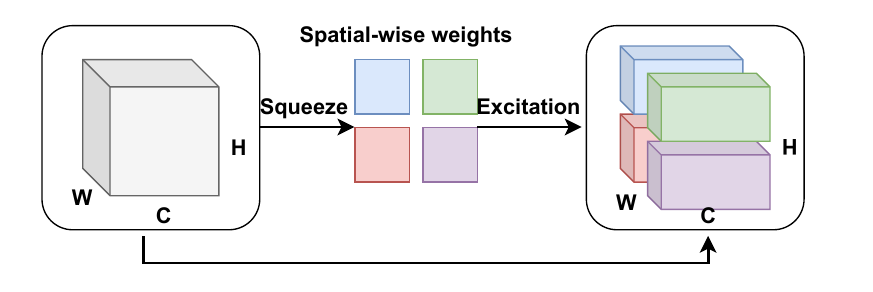}
    \caption{Spatial Attention \cite{roy2018concurrent,chen2024ehoa}}
    \label{fig:Spatial Attention}
    \vspace{-3mm}
\end{figure}

The adopted SA module structure \cite{roy2018concurrent,chen2024ehoa} is shown in Fig. \ref{fig:Spatial Attention}. 
Average pooling and max pooling are computed on the input features along the channel dimension during the squeeze stage to capture both global and salient spatial information.
% Average pooling and max pooling are computed on the input features along the channel dimension to capture spatial information. 
After that, local information extraction is performed to capture spatial relationships within the region. The attention map is then generated and normalized to ensure clear interpretability of attention weights, allowing effective modulation of features at each spatial location in subsequent steps.
The detailed implementation of spatial attention module is summarized in Algorithm \ref{alg:spatial_attention}.
\begin{algorithm}[htp]
    \small
    \caption{Spatial Attention Module}
    \label{alg:spatial_attention}
    \begin{algorithmic}[1] 
        \State \textbf{Input:} $X \in \mathbb{R}^{B \times C \times T \times H \times W}$
        \State \textbf{Output:} $\hat{X} \in \mathbb{R}^{B \times C \times T \times H \times W}$
        \State $avg\_out \gets \text{Mean}(X, \text{dim} = C, \text{keepdim} = \text{True})$
        \State $max\_out \gets \text{Max}(X, \text{dim} = C, \text{keepdim} = \text{True})$
        \State $pooled \gets \text{Concat}(avg\_out, max\_out, \text{dim} = 1)$
        \State $pooled \gets \text{Reshape}(pooled, [B , 2 , T , H, W])$
        \State $attn \gets \sigma(\text{Conv}(pooled))$
        \State $Z \gets \text{Reshape}(attn, [B , 1 , T , H, W])\cdot X$

        \State $\hat{X} \gets Z + X$
        \State \Return $\hat{X}$
    \end{algorithmic}
\end{algorithm}

Spatial-temporal Joint Attention Module:
TA and SA are combined to form a spatial-temporal joint attention module:
\begin{equation}
Y = SA\bigl(TA(X)\bigr) + X
\end{equation}

The spatial-temporal joint attention module fully utilizes temporal information before emphasizing spatial features. A residual connection preserves the expressiveness of the original features. Overall, the spatial-temporal joint attention module not only boosts event temporal modeling but also boosts event spatial modeling, enhancing facial expression recognition performance.

\section{EXPERIMENTS}  

\subsection{Dataset Preprocessing}
To validate the effectiveness and advantages of our proposed CS3D framework in facial expression recognition compared to standard baselines, mainstream datasets (ADFES \cite{van2011moving}, CASME II \cite{yan2014casme}, SZU-EmoDage \cite{han2023chinese}) were converted to event stream. To ensure consistency and robustness in the subsequent processing, each video underwent a standardized preprocessing pipeline. Specifically, facial landmark information was used to crop the facial region, accurately localizing the region of interest. Then, a rotation operation was applied to align the facial pose and eliminate bias caused by head tilt. The image was subsequently converted to grayscale to reduce the influence of color and highlight structural features. Finally, the image resolution was resized to 112×112 to meet the input size requirements of the model, reduce computational complexity, accelerate inference, and improve training stability and generalization while preserving key structural information. After completing the above normalization process, the videos were converted into event stream using the V2E converter \cite{hu2021v2e}, simulating the output of an event camera in real-world scenarios and extracting temporally dynamic event information as input for the downstream model.
% Each video is normalized through facial landmark-based cropping, rotation, grayscale conversion, and resizing to 112×112 resolution. Normalized videos are processed into event frames via V2E \cite{hu2021v2e}. 

\subsection{Model Complexity and Energy Consumption Evaluation}
% In the experiment, we focused on the computational complexity and energy consumption of our algorithm in practical applications. In the end, we proposed an evaluation method based on the THOP tool to quantitatively measure the model’s FLOPs (floating point operations) and parameter counted for a given input and further estimated its actual runtime energy consumption by incorporating the characteristics of the algorithm. The specific evaluation mechanism was described as follows:

In the experiment, we focused on the compcountedutational complexity and energy consumption of our algorithm in practical applications. We proposed an evaluation method based on the THOP tool to quantitatively measure the model’s floating point operations (FLOPs) and parameter count for a given input and further calculated its actual runtime energy consumption by incorporating the characteristics of the algorithm. The specific evaluation mechanism was described as follows:

\subsubsection{FLOPs and Parameter Count Statistics} 
${\mathrm{FLOPs}}$ represented the number of floating point operations performed by the model during inference or training, serving as an important indicator of the model’s computational complexity. Generally, a higher ${\mathrm{FLOPs}}$  count implied greater computational demand. We first constructed a tensor from event stream data to calculate the model's ${\mathrm{FLOPs}}$ and parameter count. Using the THOP tool, we recorded the number of ${\mathrm{FLOPs}}$ and parameters during the forward pass. Here, ${\mathrm{FLOPs}}$ denoted the total theoretical number of floating point operations required by the model in one forward computation. Since the directly obtained ${\mathrm{FLOPs}}$ value was typically large, we converted it into units of G (i.e., $10^9$ ${\mathrm{FLOPs}}$).

% OPs (G) denote the computational capacity in billions of operations. To compute this value, we first count the total number of operations \(N\) performed by the system during a given task. Next, by precisely measuring the total time \(T\) (in seconds) taken to complete the task, we calculate the average operation rate, \(N/T\). Finally, this rate is converted into billions of operations using the following formula:
%     %OPs (G) denoted the computational capability expressed in billions of operations. The calculation process first involved counting the total number of operations \(N\) performed by the system during a specific task; then, by accurately measuring the total time \(T\) (in seconds) taken to complete the task, the system's average operation rate, i.e. \(N/T\), was computed. For clarity, this value was subsequently converted into units of billions of operations using the following formula:
% \begin{equation}
% \text{OPs (G)} = \frac{N}{T \times 10^9}
% \end{equation}

\subsubsection{Energy Consumption Estimation Calculations}
Calculating energy consumption was crucial for the design and implementation of a real-time facial expression recognition algorithm. It helped evaluate the efficiency of the model on energy-constrained platforms and provided a basis for optimizing model architectures. By comparing the energy consumption of different algorithms, we were able to find a balance between performance and energy usage, achieving green, low-carbon computing systems that promoted energy conservation, emission reduction, and sustainable development.
% For energy consumption estimation, we further assume that each floating-point operation (FLOP) consume approximately 1 nJ of energy. Although this assumption was somewhat simplified, it served as a reasonable estimation basis in the absence of more precise hardware measurement data. Therefore, taking the pulse factor into account, the actual number of ${\mathrm{FLOPs}}$ executed by the model is: ${\mathrm{FLOPs}}_{\text{effective}} = {\mathrm{FLOPs}} \times \alpha$.
% \begin{equation}
% \text{FLOPs}_{\text{effective}} = \text{FLOPs} \times \alpha
% \end{equation}

% The corresponding total energy consumption in millijoules (mJ) is then: $E_{\mathrm{mJ}} = \mathrm{FLOPs} \times \alpha \times 10^{-6}\,\mathrm{mJ}$.
To evaluate the energy consumption performance of the model on real devices, this paper adopted a system-level measurement approach by accessing the energy monitoring interfaces provided by the devices to obtain real-time energy data during model execution. On embedded platforms such as the Jetson Xavier NX and Jetson Nano, current and voltage readings were collected from built-in energy sensors and combined with timestamps to calculate energy consumption. 
On desktop GPU platforms like the NVIDIA Titan X, real-time energy readings were collected using the nvidia-smi tool. Then the energy consumed can be roughly estimatied by:

\begin{equation}
E = \int_0^T P(t) \, dt \approx \sum_{i=1}^{N} P(t_i) \cdot \Delta t
\end{equation}

\noindent
where $\Delta t$ is the fixed sampling interval, $N$ is the number of samples. $P(t_i)$ denotes the instantaneous power at the $i$-th sampling point, which can be obtained by the collected information on different devices introduced previously.
% with real-time energy readings using the \textit{nvidia-smi} tool.
 
% \begin{equation}
% E = \mathrm{FLOPs}_{\mathrm{effective}} \times 1\,\text{nJ}
% = \mathrm{FLOPs} \times \alpha \times 10^{-9}\,\text{J}
% \end{equation}
%
% For ease of presentation, we convert energy consumption to millijoules (mJ):
% \begin{equation}
% E_{\mathrm{mJ}} = \mathrm{FLOPs} \times \alpha \times 10^{-6}\,\mathrm{mJ}
% \end{equation}

% This method provided a straightforward and intuitive approach for evaluating a model’s energy overhead in real-world deployments. Although it involved some simplified assumptions, it still served as an important reference when comparing the energy consumption performance of different models. Finally, the following comparison between the C3D method and the CS3D method was obtained as TABLE Ⅰ:
This approach offered a clear and intuitive way to assess the energy overhead of a model in real-world deployments. Although it relied on some simplified assumptions, it served as a valuable benchmark to compare the energy consumption of different models. Table I presented a comparative analysis of the energy consumption of the C3D and CS3D methods on different computing devices.
\begin{table}[ht]
\centering
\small 
\setlength{\tabcolsep}{5pt} 
\renewcommand{\arraystretch}{1.3} 
\caption{Energy Consumption Comparison of C3D and CS3D Architecture on Different Computing Devices.}
\begin{tabular}{lccc}
\hline\hline
\textbf{Method} & \textbf{Platform} & \textbf{FLOPs (G)} & \textbf{Energy (mJ)} \\
\hline
     & Jetson Nano       & 21.29 & $3.71 \times 10^3$ \\
C3D  & Jetson Xavier NX  & 21.29 & $25.7 \times 10^3$ \\
     & Titan X           & 21.29 & $18.2 \times 10^3$ \\
\hline
     & Jetson Nano       & 4.68  & $10.3 \times 10^3$ \\
CS3D (Ours) & Jetson Xavier NX  & 4.68  & $6.74 \times 10^3$ \\
     & Titan X           & 4.68  & $4.01 \times 10^3$ \\
\hline\hline
\end{tabular}
\label{tab:energy_comparison}
\end{table}

\subsection{Accuracy Rate Comparison}
% In this experiment, we compared the performance of different methods on the ADFES, CASME II, and SZU-EmoDage datasets, including RNN \cite{sherstinsky2020fundamentals}, Transformer \cite{vaswani2017attention}, LSTM, C3D, and our proposed CS3D method. 
In this experiment, we compared the performance of different algorithms on the ADFES, CASME II, and SZU-EmoDage datasets, including RNN \cite{sherstinsky2020fundamentals}, Transformer \cite{vaswani2017attention}, LSTM, C3D, and our proposed CS3D method. For fair comparison, all baseline models were implemented using standard configurations: the RNN and LSTM models consist of two recurrent layers with 128 hidden units each, followed by a fully connected classification layer; the Transformer model includes 2 encoder layers with 4 attention heads and a model dimension of 256; the traditional C3D model follows the original design with five 3D convolutional layers and two fully connected layers. All algorithms were trained using the Adam optimizer with a learning rate of 1e-4 and a batch size of 16.
\begin{table}[h]
\caption{Comparison results of rate across different models and datasets.}
\centering
\renewcommand{\arraystretch}{1.2} 
\setlength{\tabcolsep}{1pt} 
\small 
\begin{tabular}{l@{}c@{}c@{}c@{}c@{}c}
    \hline\hline
    Dataset & RNN & Transformer & LSTM & C3D  & CS3D(Ours)  \\ 
    \hline
    ADFES & 40.54\% & 51.35\% & 29.73\% & 70.27\% & \textbf{78.38}\% \\ 
    CASME II & 36.73\% & 42.86\% & 40.82\% & 40.82\% & \textbf{54.79}\% \\ 
    SZU-EmoDage & 33.33\% & 28.03\% & 29.55\% & 79.45\% & \textbf{90.91\%} \\ 
    \hline\hline
\end{tabular}
\vspace{-3mm}
\label{tab:ComparisonAccuracy}
\end{table}
% \begin{table}[t]
% \caption{Comparison of accuracy rate across models and datasets.}
% \label{tab:ComparisonAccuracy}
% \centering
% \footnotesize
% \setlength{\tabcolsep}{3pt}
% \renewcommand{\arraystretch}{1.1}
% \resizebox{\linewidth}{!}{%
% \begin{tabular}{lccccc}
% \specialrule 
% Dataset & RNN & Transformer & LSTM & C3D & CS3D (ours) \\
% \midrule
% ADFES        & 40.54\% & 51.35\% & 29.73\% & 70.27\% & \textbf{78.38}\% \\
% CASME II     & 36.73\% & 42.86\% & 40.82\% & 40.82\% & \textbf{54.79}\% \\
% SZU-EmoDage  & 33.33\% & 28.03\% & 29.55\% & 79.45\% & \textbf{90.91}\% \\
% \specialrule 
% \end{tabular}%
% }
% \end{table}

\begin{figure}[b!]
    \centering
    \includegraphics[width=1\linewidth]{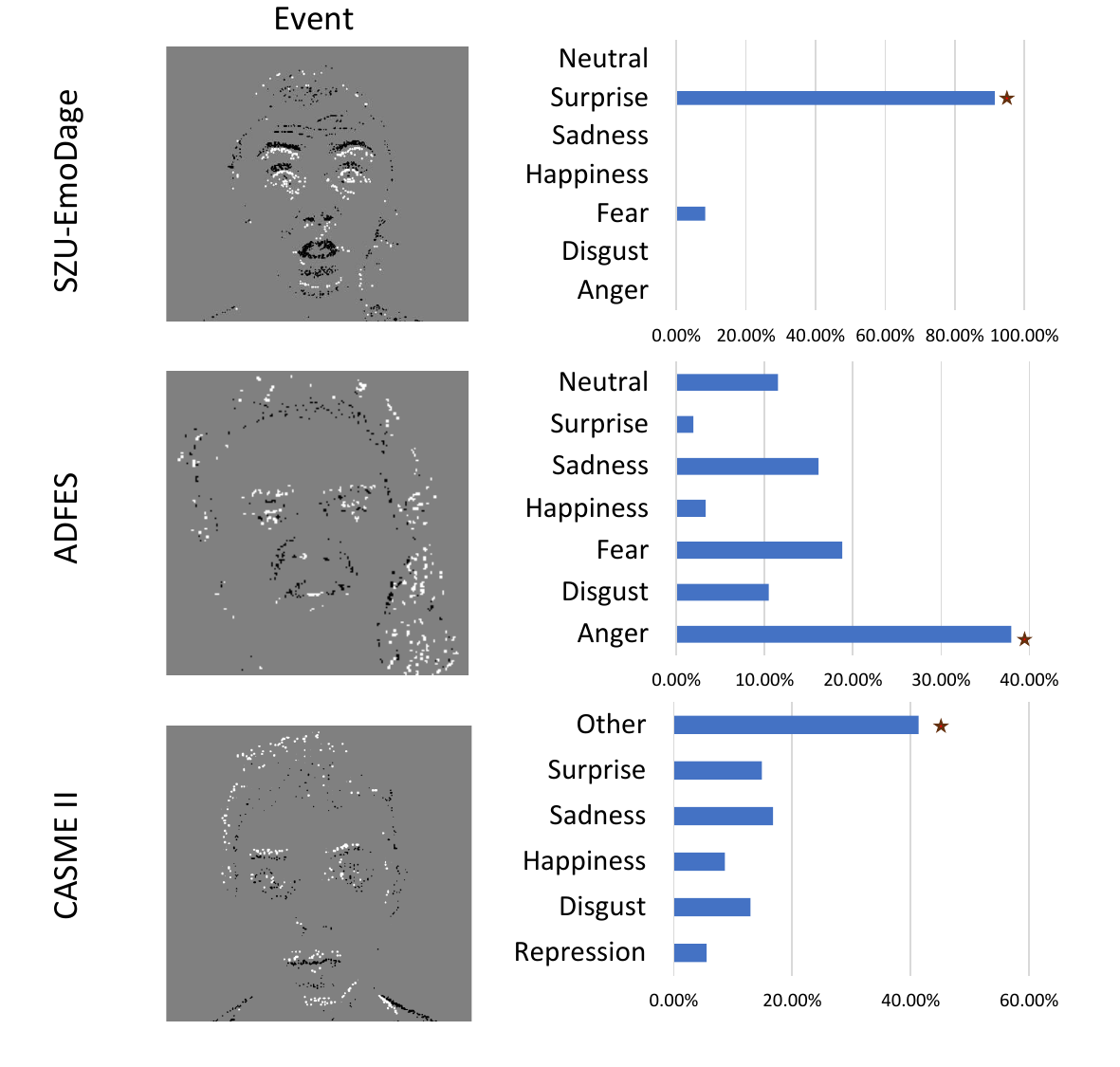}
    \vspace{-4mm}
    \caption{Visualization of the raw event streams and the output results of our CS3D method, demonstrating three emotion tasks: Surprise (SZU-EmoDage), Anger (ADFES), and Others (CASME II).}
    \label{fig:enter-label}
\end{figure}

The experimental results, as shown in Table~\ref{tab:ComparisonAccuracy}, indicate that the LSTM module exhibited the lowest accuracy rate on the ADFES dataset, achieving a mere 29.73\%. Similarly, the RNN algorithm recorded its poorest performance on the CASME II dataset with an accuracy rate of 36.73\%, while the Transformer algorithm demonstrated the least efficacy on the SZU-EmoDage dataset, attaining only 28.03\% accuracy. In contrast, our proposed CS3D framework achieved the best performance across all datasets, reaching 78.38\% on ADFES, 54.79\% on CASME II, and 90.91\% on SZU-EmoDage. Compared to the traditional C3D algorithm (which achieved 70.27\%, 40.82\%, and 78.38\% in the respective datasets), CS3D showed significant improvements. These results indicate that the CS3D method can more effectively capture spatial-temporal features in event stream, thereby enhancing video-based facial expression recognition performance.
The Transformer underperforms on the SZU-EmoDage dataset because it contains subtle and temporally continuous facial motions that demand strong temporal modeling. LSTMs and RNNs better capture such sequential dependencies, while the Transformer’s attention mechanism struggles with limited event-based data. 
In contrast, on datasets dominated by spatial cues and global correlations, the Transformer performs better, showing that its effectiveness depends on temporal dynamics and data scale.

% \subsection{Accuracy Comparison Experiments}
% To further investigate the specific experimental results of the expression prediction accuracy, 
Furthermore, to provide a more intuitive demonstration of the experimental effectiveness of the CS3D architecture, we evaluated the trained module on the SZU-EmoDage, ADFES, and CASME II datasets for emotion recognition and conducted a comprehensive performance comparison, as shown in Fig.~\ref{fig:enter-label}.   The results indicate that the algorithm achieves accuracy rates of 91.45\%, 37.95\%, and 41.30\% for “Surprise,” “Anger,” and “Other” expressions, respectively, demonstrating accuracy rate and robust performance for facial expression recognition. 

% [91.45\% accuracy for surprised expressions, 37.95\% accuracy for angry expressions, and 41.30\% accuracy for "other" expressions respectively, demonstrating robust performance for event camera-based recognition.

% \begin{figure}[htp]
%     \centering
%     \includegraphics[width=1\linewidth]{experiment.pdf}
%     \caption{Visualization of the raw event streams and the output results of our CS3D method, demonstrating three emotion tasks: Surprise (SZU-EmoDage), Anger (ADFES), and Others (CASME II)}
%     \label{fig:enter-label}
% \end{figure}

To further validate the superiority of our approach in multi-class recognition scenarios, we specifically benchmarked the proposed method against state-of-the-art approaches on the ADFES dataset using its challenging seven-class classification task. Notably, this dataset remains under-explored for fine-grained emotion categorization, with limited existing studies addressing its full seven-class recognition potential. As shown in Table III, our method achieves a significant performance advantage over Spiking-Fer\cite{barchid2023spiking}, including their enhanced variants with data augmentation techniques. 

%We conducted experiments on the dataset ADFES to validate the effectiveness of our method. Few studies have performed seven-class classification on the ADFES dataset, so we replicated the methods from two papers and presented the results in Table III. In the seven-class facial expression recognition task, our method achieved the best results, achieving higher accuracy than MMNet\cite{li2022mmnet} , Spiking-Fer\cite{barchid2023spiking} , and various Spiking-Fer based data augmentation methods.

\begin{table}[htp]
    \caption{ Comparison results on the ADFES dataset. 
    % (A) Best configuration based on common event data augmentations; (B) With the addition of eventdrop; (C) With the addition of eventdrop and mirror.
    }
    \label{tab:accuracy_comparison}
    \centering
    \begin{tabular}{lc}
        \hline\hline
        Method & Accuracy  \\
        \hline
        Spiking-Fer \cite{barchid2023spiking} & 47.00\% \\
        Spiking-Fer + A  \cite{barchid2023spiking}& 60.40\% \\
        Spiking-Fer + B \cite{barchid2023spiking} & 61.50\% \\
        Spiking-Fer + C \cite{barchid2023spiking} & 74.20\% \\
        % MMNet \cite{li2022mmnet}  & 72.97\% \\
        CS3D (ours) & \textbf{78.38\%} \\
        \hline\hline \\ [0.1mm]
    \end{tabular} \\
    \vspace{-2mm}
    \begin{tablenotes}
        \item Note: A, B, and C are the event vision data augmentation methods adopted in \cite{barchid2023spiking}. A refers to ``Best configuration based on common event data augmentations", B refers to ``With the addition of eventdrop", C refers to ``With the addition of eventdrop and mirror".
    \end{tablenotes}
    
\end{table}

\subsection{Real World Validation}
\begin{figure}[h!]
    \centering
    \includegraphics[width=1\linewidth]{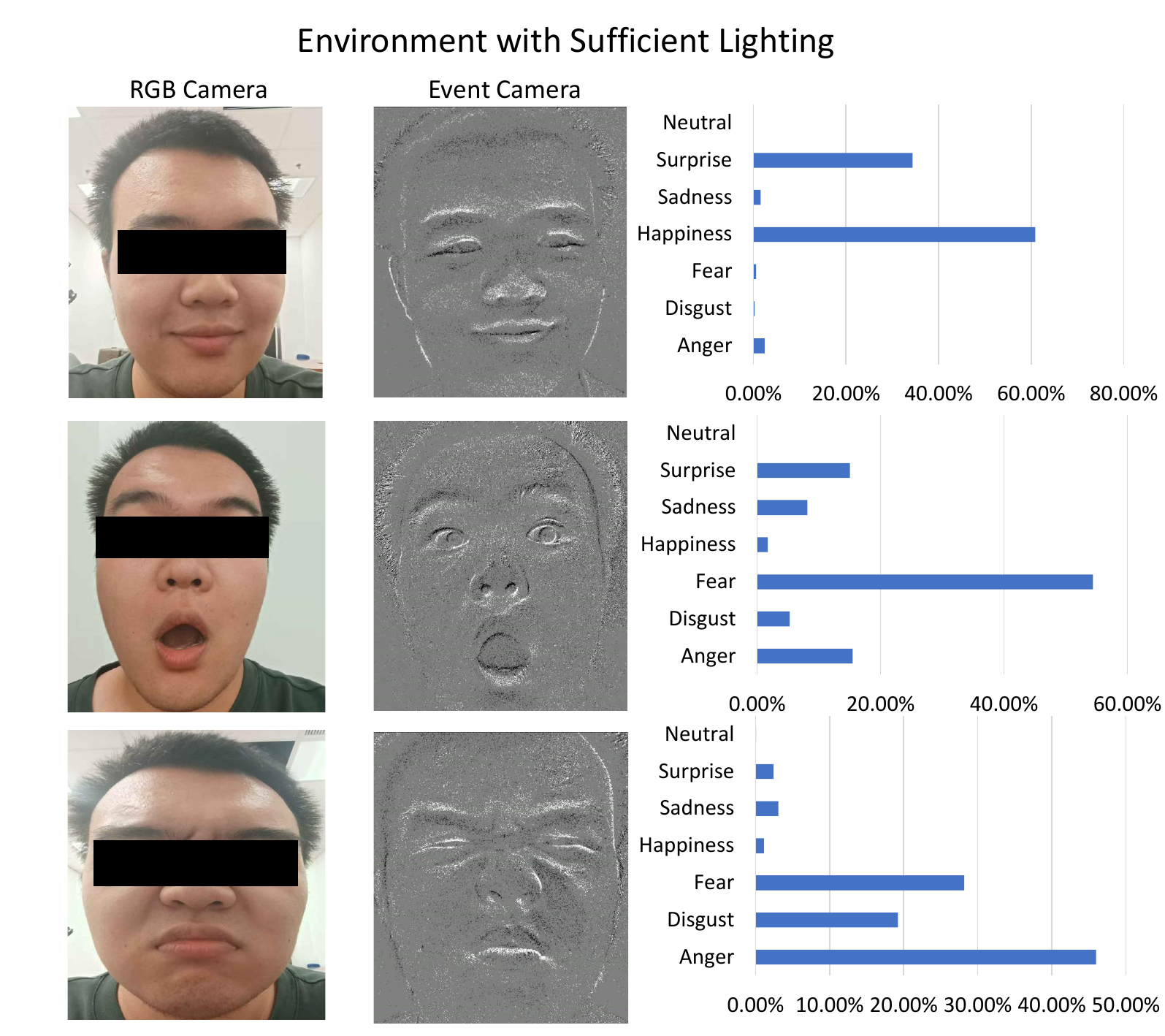}
    \caption{The event camera performs facial expression recognition in a sufficient light environment.}
    \label{fig:RealDetectionHighlight}
\end{figure}

% To further validate the superiority of our method in multi-class recognition scenarios, we perform real-time facial expression recognition under varying lighting conditions using an event camera. The visualization of the results is shown in Fig. A.
To further validate the superiority of the proposed method in multi-class facial expression recognition scenarios, real-time experiments were conducted under different lighting conditions using an event camera. In a sufficient lighting environment, as shown in Fig. \ref{fig:RealDetectionHighlight}, the participants performed typical facial expressions such as happiness, anger, and fear, which were accurately recognized by the proposed algorithm. The results demonstrate that, in environments with sufficient lighting, the event camera can capture finer details of facial muscle movements, thereby improving the recognition accuracy rate.

\begin{figure}[h!]
    \centering
    \includegraphics[width=1\linewidth]{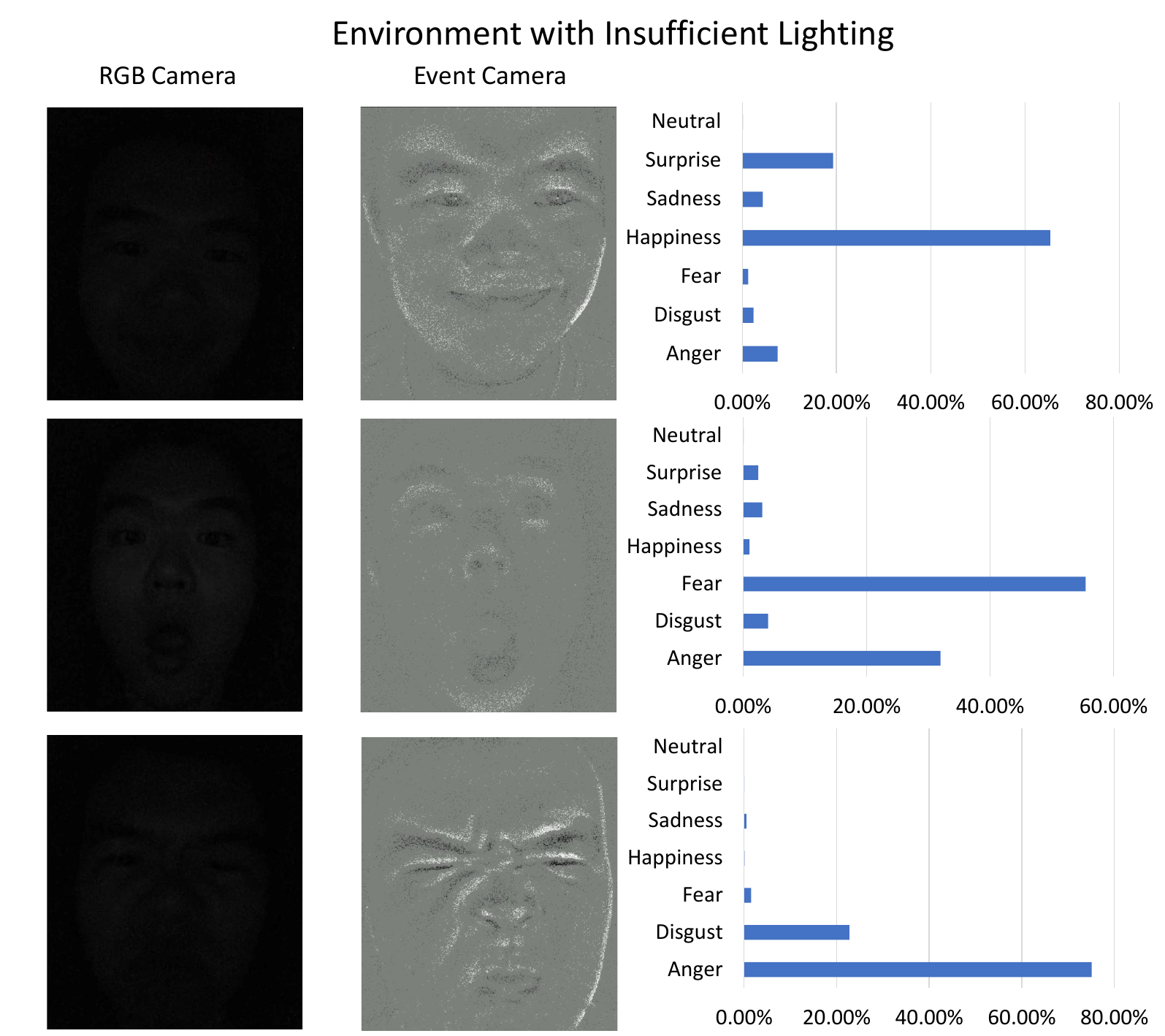}
    \caption{The event camera performs facial expression recognition in an insufficient light environment.}
    \label{fig:RealDetectionLowlight}
\end{figure}

In contrast, the results under the insufficient lighting environment were presented in Fig. \ref{fig:RealDetectionLowlight}. Compared to RGB cameras, which struggled to extract facial information in dim environments due to limited imaging capability, the event camera maintained stable performance because of its high dynamic range and motion blur-free characteristics. It continued to capture reliable event data for accurate expression recognition. These findings confirmed that the proposed method remained robust and effective even under extreme or challenging lighting conditions.

\begin{table}[h]
    \caption{Comparative analysis of accuracy, computational complexity, and energy consumption on the SZU-EmoDage dataset using the Titan X GPU.}
    \label{tab:ablation study}
    \centering
    \scriptsize  
    \begin{tabular}{lccc}  
        \hline\hline
        Method & Accuracy & FLOPs(G) & Energy(mJ) \\
        \hline
        C3D & 79.45\% & $21.29$ & $1.82\times 10^4$\\
        C3D + SSNs & 83.73\% & $21.24$ & $1.82\times 10^4$\\
        C3D + FactorizedConv3D & 87.88\% & \textbf{4.26} & $\bm{3.65 \times 10^3}$\\
        C3D + Spatial-temporal Attention & 90.17\% & $21.48$ & $1.83\times 10^4$\\
        CS3D \textbf{(ours)} & \textbf{90.91\%} & $\underline{4.68}$ & $\underline{4.01\times 10^3}$\\[0.5mm]
        \hline\hline
    \end{tabular}
    % \vspace{-3mm}
\end{table}

\subsection{Ablation Study}
Table \ref{tab:ablation study} showed an ablation study for facial expression recognition conducted on Titan X GPU. We evaluated accuracy rate, computational complexity, and energy consumption. The baseline C3D method achieved a accuracy rate of 79. 45\% with a computational complexity of 21.29G and a consumption of \(1.82 \times 10^4\) mJ of energy. The incorporation of SSNs increased accuracy to 83.73\% but did not significantly reduce complexity or energy. The addition of factorized 3D convolution raised the accuracy rate to 87. 88\% while reducing computational complexity to 4.26 G and energy consumption to \(3.65 \times 10^3\) mJ. The inclusion of a spatial-temporal joint attention mechanism further increased accuracy rate to 90.17\%, although it slightly raised computational complexity and energy consumption. Finally, the proposed CS3D method achieved the highest accuracy of 90.91\% with 4.68 G computational complexity and \(4.01 \times 10^3\) mJ energy consumption. These results indicate that CS3D not only achieves the highest recognition accuracy but also significantly reduces computational complexity and energy consumption, reaching a near-optimal level of energy efficiency. Compared to the original C3D network, it demonstrates a substantial reduction in energy usage, making it well-suited for deployment on edge computing devices to reduce the overall cost of robotic applications.

\section{CONCLUSIONS}

% In order to achieve reliable face detection on edge computing devices for service robots and to sustain stable, accurate performance in low-light scenarios.
% FER on edge devices remains challenging, as existing methods often face low recognition accuracy rate, low-light limitations, and high energy consumption.
% Event-based FER is hard to deploy on edge devicescomplex, power-hungry, and inaccurate in low light.
The event camera is becoming more widely
utilized in capturing dynamic and subtle changes due to its high temporal resolution, low latency, computational efficiency, and robustness in low-light conditions. However, event-based FER is challenging as existing methods still suffer from inaccurate and energy-intensive limitations, especially when deploying on edge computing devices. Consequently, this work proposes an efficient CS3D framework for event-based FER, which integrates soft spiking neurons, a factorized 3D convolution module, and a spatial-temporal joint attention mechanism. 
% The proposed method fully exploits the spatio-temporal dynamic features of facial expressions in event stream and employs an adaptive attention mechanism to highlight key expression information, thereby efficiently capturing subtle expression changes. 
% By \redcolor{and thus reducing the computatiobal costs and energy consumption }.
% To reduce energy consumption, we propose factorized 3D convolution by decomposing 3D convolutional to reduce parameters and energy consumption.
% Experimental results demonstrated that our proposed CS3D framework achieves high precision, robustness, and effectiveness for FER. 
Compared to the traditional C3D directly implemented for event-based FER, the proposed CS3D framework improves the accuracy rate by 8.11\% on the ADFES dataset, 13.97\% on CASME II, and 11.46\% on SZU-EmoDage. Moreover, CS3D reduces energy consumption to only 21.97\% of the C3D framework on the Titan X GPU when evaluated on the SZU-EmoDage dataset. 
% Furthermore, we evaluate CS3D for FER using an event camera in both sufficient and insufficient light environments, and it operates reliably in both cases. 
The conducted experiments indicate that our proposed CS3D framework for event-based FER achieves high efficiency with low energy consumption, high accuracy, and good robustness as it still operates reliably even under insufficient lighting conditions. In the future, integrating other modalities, such as audio, text, or physiological signals, would help to capture more emotional information and enhance the recognition of complex emotions and subtle changes.


%%%%%%%%%%%%%%%%%%%%%%%%%%%%%%%%%%%%%%%%%%%%%%%%%%%%%%%%%%%%%%%%%%%%%%%%%%%%%%%%

%%%%%%%%%%%%%%%%%%%%%%%%%%%%%%%%%%%%%%%%%%%%%%%%%%%%%%%%%%%%%%%%%%%%%%%%%%%%%%%%

%%%%%%%%%%%%%%%%%%%%%%%%%%%%%%%%%%%%%%%%%%%%%%%%%%%%%%%%%%%%%%%%%%%%%%%%%%%%%%%%
% \clearpage
\bibliographystyle{IEEEtran}
\normalem
% \balance
\bibliography{IEEEabrv}

\end{document}